\documentclass[lettersize,journal]{IEEEtran}
\usepackage[utf8]{inputenc}
\usepackage{array}
\usepackage{mdwmath}
\usepackage{mdwtab}
\usepackage{eqparbox}
\usepackage{url}
\usepackage{cite}
\usepackage{amsmath,amssymb,amsfonts}
\usepackage{algorithmic}
\usepackage{textcomp}
\usepackage{lettrine}
\usepackage{caption}
\usepackage{subcaption}
\usepackage[export]{adjustbox}
\usepackage{amsmath}
\newcommand{\subparagraph}{}
\usepackage{leftidx}
\usepackage{siunitx}
\usepackage{multirow}
\usepackage{balance}
\usepackage[export]{adjustbox}
\usepackage{booktabs}
\usepackage{bbold}

\DeclareMathOperator*{\argmin}{arg\,min}
\DeclareMathOperator*{\atantwo}{atan2}
\newcommand{\rulesep}{\unskip\ \vrule\ }

\begin{document}
\markboth{IEEE Robotics and Automation Letters. Preprint Version. Accepted June, 2022}
{Bavle \MakeLowercase{\textit{et al.}}: Situational Graphs} 

\title{\LARGE \bf Situational Graphs for Robot Navigation \\ in Structured Indoor Environments 
}

\author{Hriday Bavle$^{1}$, Jose Luis Sanchez-Lopez$^{1}$, Muhammad Shaheer$^{1}$, \\ Javier Civera$^{2}$ and Holger Voos$^{1}$ 
\thanks{Manuscript received: Feb, 24, 2022; Revised May, 26, 2022; Accepted June, 23, 2022.} 
\thanks{This paper was recommended for publication by Editor Sven Behnke upon evaluation of the Associate Editor and
Reviewers’ comments.} 
\thanks{*This work was partially funded by the Fonds National de la Recherche of Luxembourg (FNR), under the projects C19/IS/13713801/5G-Sky, by a partnership between the Interdisciplinary Center for Security Reliability and Trust (SnT) of the University of Luxembourg and Stugalux Construction S.A., by the Spanish Government under Grants PGC2018-096367-B-I00 and PID2021-127685NB-I00 and by the Arag{\'o}n Government under Grant DGA T45 17R/FSE.
For the purpose of Open Access, the author has applied a CC BY public
copyright license to any Author Accepted Manuscript version arising from
this submission.}
\thanks{$^{1}$Authors are with the Automation and Robotics Research Group, Interdisciplinary Centre for Security, Reliability and Trust, University of Luxembourg. Holger Voos is also associated with the Faculty of Science, Technology and Medicine, University of Luxembourg, Luxembourg.
\tt{\small{\{hriday.bavle, joseluis.sanchezlopez, muhammad.shaheer, holger.voos\}}@uni.lu}}%
\thanks{$^{2}$Author is with I3A, Universidad de Zaragoza, Spain
{\tt\small jcivera@unizar.es}}%
\thanks{Digital Object Identifier (DOI): see top of this page.}
}

\maketitle

\begin{abstract}
Mobile robots should be aware of their \emph{situation}, comprising the deep understanding of their surrounding environment along with the estimation of its own state, to successfully make intelligent decisions and execute tasks autonomously in real environments.
3D scene graphs are an emerging field of research that propose to represent the environment in a joint model comprising geometric, semantic and relational/topological dimensions. Although 3D scene graphs have already been combined with SLAM techniques to provide robots with situational understanding, further research is still required to effectively deploy them on-board mobile robots.

To this end, we present in this paper a novel, real-time, online built Situational Graph (\textit{S-Graph}), which combines in a single optimizable graph, the representation of the environment with the aforementioned three dimensions, together with the robot pose. 
Our method utilizes odometry readings and planar surfaces extracted from 3D LiDAR scans, to construct and optimize in real-time a three layered \textit{S-Graph} that includes (1) a robot tracking layer where the robot poses are registered, (2) a metric-semantic layer with features such as planar walls and (3) our novel topological layer constraining the planar walls using higher-level features such as corridors and rooms. 
Our proposal does not only demonstrate state-of-the-art results for pose estimation of the robot, but also contributes with a metric-semantic-topological model of the environment.


\end{abstract}
\section{Introduction}

\IEEEPARstart{M}{obile} robots require rich semantic descriptions of the key elements of a scene to understand the situation around them, to estimate accurate task-oriented maps of their surrounding environment and to localize themselves into them. Geometric LiDAR SLAM methods such as \cite{loam}, \cite{cartographer} and \cite{hdl_graph_slam} are essential for safe navigation but they are unable to identify these key elements, which compromises among others high-level task definitions, the localization and mapping performance in cluttered, repetitive or dynamic environments. Semantic SLAM methods like \cite{probablisitic_bowman}, \cite{vps_slam} and \cite{suma++} utilize key semantic elements from the environment but do neither consider nor model the relation between these elements to further constraint their geometry, which would improve their performance and the understanding of the situation around the robot. 

\begin{figure}[t]
    \centering
    \includegraphics[width=0.5\textwidth]{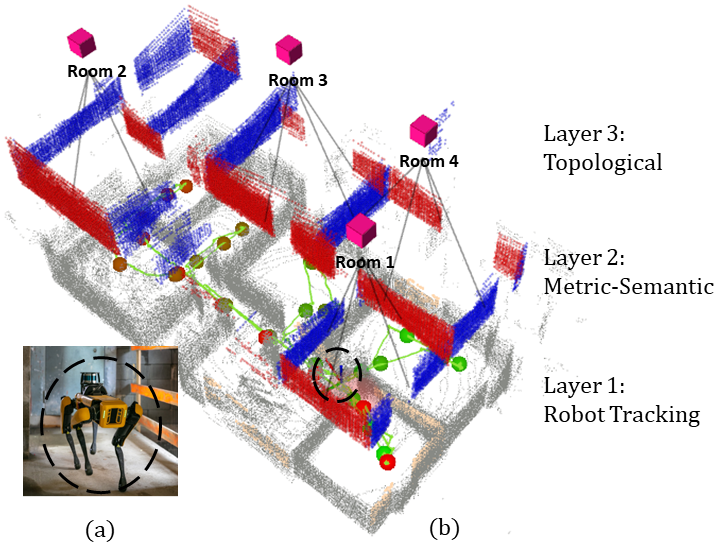}
    \caption{\textit{S-Graph} generated using the legged robot in (a) corresponding to a real construction site. The graph in (b) can be divided into three sub-layers: 1) The robot tracking layer estimates the sensor poses and creates a map of keyframes. 2) The metric-semantic layer creates a plane-based map connected with the keyframes. 3) The topological layer links the planes and the room/corridor representations using novel factors proposed on this work.}
    \label{fig:scene_graph}
\end{figure}

\begin{figure*}[!ht]
    \centering
    \includegraphics[width=1\textwidth]{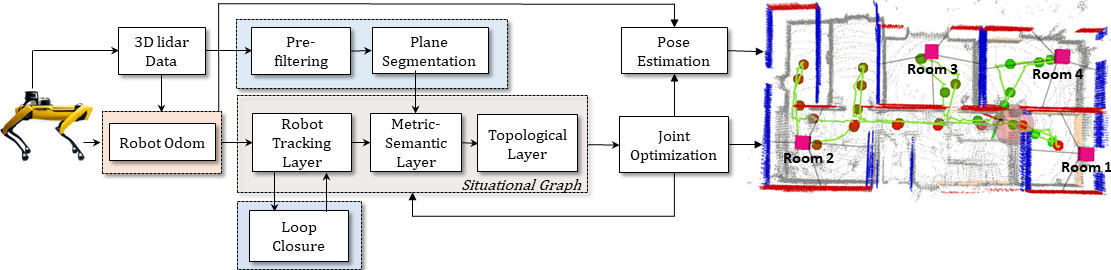}
    \caption{Pipeline of the proposed \textit{S-Graph} architecture, receiving 3D LiDAR measurements and robot odometry at a given time instant $t$, along with its filtering and planar extraction components. The figure also presents the three layered situational graph along with the loop closure constraint, which are jointly optimized to update the robot pose and the \textit{S-Graph}.}
    \label{fig:system_architecture}
\end{figure*}

Recent approaches such as \cite{3d_scene_graph}, \cite{dynamic_scene_graph}, \cite{scene_graph_fusion} model the scene as a graph, in order to efficiently represent the environment and its semantic elements in a hierarchical representation with structural and topological constraints between the elements. Scene graphs enable the robots to understand and navigate the environment similarly to humans, using high-level abstractions (such as chairs, tables, walls) and the inter-connections between them (such as a set of walls forming a room or a corridor). Although works using scene graphs show promising results, there still exists a gap to connect such scene graphs with SLAM methods that simultaneously optimize the robot poses along with the scene graph. 

In this direction we present Situational Graphs (\textit{S-Graphs}), which bridges geometric LiDAR SLAM and scene graphs. Our \textit{S-Graphs} (Fig.~\ref{fig:scene_graph}) are divided into three layers, namely \textit{Robot Tracking Layer}, \textit{Metric-Semantic Layer} and \textit{Topological Layer}. The robot tracking layer is the lowest level of the graph creating keyframes at regular time-distance intervals whose pose is constrained by odometry measurements. The metric-semantic layer extracts semantic elements at a given keyframe and maps them using their geometric constraints with respect to the keyframes. In our case we use pose-plane constraints between detected planar surfaces and keyframes.
The topological layer is the highest level of the graph connecting the mapped planar surfaces to higher-level entities using suitable topological constraints, which in our case are the room and corridor factors. These factors add additional constraints over their corresponding plane features. In this manner, \textit{S-Graphs} do not only optimize the robot poses but also the whole 3D scene graph jointly, comprising the mapped planar surfaces along with the corridor and room geometries. 
Our main contributions in this paper are: 

\begin{itemize}
    \item A real-time implementation of \textit{S-Graphs} using robot odometry and 3D LiDAR measurements with three hierarchical layers, optimizing the robot poses jointly with a high-level 3D representation of the scene.   
    \item The introduction of novel room and corridor factors into a graph formulation, that enable high-level representations of indoor scenes and constrain planar landmarks and robot poses. 
    \item A thorough experimental evaluation on simulated and real scenarios, showing that in addition to representation advantages, our \textit{S-Graph} model achieves state-of-the-art accuracy. 
\end{itemize}

\section{Related Works}

\subsection{Geometric SLAM} 
Geometric SLAM, using either visual sensors or 3D LiDAR, have been widely researched in the field of robotics in the last decades. We refer the reader to the survey in \cite{cadena2016past}, \cite{sa_survey} for a broad perspective of the research in the field, and to ORB-SLAM2 \cite{orbslam2} and LOAM \cite{loam} as two representative pipelines. Both of them estimate in real-time the global robot pose using visual/LiDAR measurements, as well as accurate 3D point clouds as map representation. 

Focusing on LiDAR SLAM, similarly to LOAM and its variants \cite{floam}, \cite{mloam}, \cite{sca_loam}, methods like LIO-SAM \cite{lio_sam}, LIO-MAPPING \cite{lio_mapping}, HDL-SLAM \cite{hdl_graph_slam} estimate the robot poses and a 3D map, with option of integrating additional sensor modalities such as IMU and GPS. Other methods such as like LIMO \cite{limo}, LIRO \cite{liro} and LVI-SLAM \cite{lvi_slam} fuse visual and LiDAR measurements for simultaneous localization and mapping. Although these systems demonstrated significant progress in robustness and accuracy in the last years, they are limited for several application cases by their map representation. The lack of high-level semantically meaningful map representations reduces their robustness and hinders their use for interfacing with humans or defining object-based tasks.

\subsection{Semantic SLAM}
To improve the robustness and richness of representation of geometric SLAM, semantic SLAM has evolved with methods such as  
\cite{probablisitic_bowman}, VPS-SLAM \cite{vps_slam}, \cite{probablistic_doherty} estimating a geometric map and adding semantic objects in the environment for jointly optimizing the robot pose and semantic object landmarks. Methods like \cite{kimera_only}, \cite{voxblox_plus} utilize object detections to create dense metric-semantic maps of the environment. Works using LiDAR sensors like LeGO-LOAM \cite{lego-loam} incorporate ground planes and line edges as high-level features from segmented point cloud data to improve robot poses and map estimates when compared against LOAM. SA-LOAM \cite{sa_loam} utilizes semantics from the environment like roads, buildings, traffic signs to improve the loop closure accuracy of LOAM. Similarly, methods like SUMA++ \cite{suma++} utilize semantics to filter out dynamic objects improving the robot pose and the map quality. SegMap \cite{segmap} presents a semantic mapping and localization solution, learning data-driven descriptors using segments extracted from 3D point clouds. 
All the above methods outperform their geometric SLAM counterparts being able to classify and map different semantic elements in the environment, but they can still suffer errors due to misidentification and errors in the pose estimate of the semantic elements. Adding structural/topological constraints between different semantic elements can further increase the robustness of the environmental understanding and reasoning for the robot. Similar to our approach, methods like $\pi-$LSAM \cite{pi_slam}, \cite{lidar_slam_with_plane_adj} utilize segmented planar features to create planar maps but as the other semantic SLAM approaches, they do not consider any topological/structural constraints between the planes.

\subsection{Scene Graphs}
Recent advances in computer vision have led to development of scene graphs, generating a comprehensive graph of all the extracted semantic information from an environment along with the inter-relationships of the different semantic components within it. The pioneering work of 3D Scene Graph \cite{3d_scene_graph} creates an offline semi-autonomous framework using object detections over RGB images, generating a multi-layered hierarchical representation of the environment and it components, divided mainly in layers of camera, objects, rooms and building. 

Rosinol et al. \cite{kimera} extend the 3D scene graph concept to environments with static and dynamic agents. It uses the Kimera Visual-Inertial Odometry (VIO) \cite{kimera_only} with object detections to create a metric-semantic mesh which is then fed to a scene generator creating the dynamic scene graph in an offline fashion. SceneGraphFusion \cite{scene_graph_fusion} on the other hand, generates a real-time incremental 3D scene graph using RGB-D sequences, accurately handling partial and missed semantic data. Though promising in terms of scene representation, a major drawback of these models is that they do not tightly couple the estimate of the scene graph with the state estimate of the SLAM framework, to simultaneously optimize them, thus they generate a scene graph and a SLAM graph in an independent manner.  
The very recent work Hydra \cite{hydra}, presents research in the direction of real-time scene graph generation as well as its optimization using loop closure constraints. Our approach is aligned in a similar direction towards real-time scene graphs. However, compared to Hydra, ours creates the entire \textit{S-Graph} as an optimizable factor graph, being constrained as a whole whenever planar landmarks are observed and specifically linking the planar landmarks with our novel topological constraints.         


\section{Situational Graphs}
\label{proposed_method}

\subsection{Overview}
An overview of the proposed approach is shown in Fig.~\ref{fig:system_architecture}. 
We define four reference frames: the map frame $M$, the odometry frame $O$, the robot frame $R_t$ and the LiDAR frame $L_t$. The last two change with time as the robot moves but are rigidly related to each other. We separate the odometry and map frames to avoid large state updates affect the navigation commands. We include in the state their relative transformation $\leftidx{^M}{\mathbf{x}}_{O}$.
Our pipeline receives as input the 3D LiDAR measurements in frame $L_t$ as well as odometry measurements from the robot sensors in frame $R_t$. Our global state $\mathbf{s}$ is defined as
\begin{multline}
\mathbf{s} = [\leftidx{^M}{\mathbf{x}}_{R_1}, \ \hdots, \ \leftidx{^M}{\mathbf{x}}_{R_T}, \ \leftidx{^M}{\boldsymbol{\pi}}_{1}, \ \hdots, \ \leftidx{^M}{\boldsymbol{\pi}}_{P}, \\
\leftidx{^M}{\boldsymbol{\rho}}_{1}, \ \hdots, \ \leftidx{^M}{\boldsymbol{\rho}}_{S}, \ \leftidx{^M}{\boldsymbol{\kappa}}_{1}, \ \hdots, \  \leftidx{^M}{\boldsymbol{\kappa}}_{K}, \leftidx{^M}{\mathbf{x}}_{O}]^\top,
\end{multline}

\noindent where $\leftidx{^M}{\mathbf{x}}_{R_t}, \ t \in \{1, \hdots, T\}$ are the robot poses at $T$ selected keyframes, $\leftidx{^M}{\boldsymbol{\pi}}_{i}, \ i \in \{1, \hdots, P\}$ are the plane parameters of the $P$ planes in the scene, $\leftidx{^M}{\boldsymbol{\rho}}_{j}, \ j \in \{1, \hdots, S\}$ contains the parameters of the $S$ rooms and $\leftidx{^M}{\boldsymbol{\kappa}}_{k}, \ k \in \{1, \hdots, K\}$ the parameters of the $k$ corridors. We detail the specific form of these map elements later in Section~\ref{subsec:s-graph}.

We will jointly optimize the state $\mathbf{s}$ as follows
\begin{equation}
    \hat{\mathbf{s}} = \argmin_{\mathbf{s}} (c_{\mathbf{x}} + c_{\boldsymbol{\pi}} +  c_{\boldsymbol{\rho}} + c_{\boldsymbol{\kappa}})
\end{equation}

\noindent where $c_{\mathbf{x}}$,  $c_{\boldsymbol{\pi}}$, $c_{\boldsymbol{\rho}}$ and $c_{\boldsymbol{\kappa}}$ are cost functions related respectively with the robot tracking, metric-semantic and topological (room and corridor) layers. We elaborate on them in Section~\ref{subsec:s-graph}.

The overall pipeline can be divided into four main modules (colored boxes in Fig.~\ref{fig:system_architecture}). The first one pre-filters the LiDAR measurements to remove noise and performs plane segmentation to detect and extract the planar surfaces. The second module computes the robot odometry either from LiDAR measurements or from the robot encoders. The third module is the LiDAR mapping, which generates the \textit{S-Graph}. It first receives the robot odometry measurements to create a factor graph of robot keyframe poses as factor nodes at pre-defined distances. These nodes form the first layer of the \textit{S-Graph}. It also receives the plane detections for the corresponding keyframes, which are classified either as vertical or horizontal planar nodes, forming in this manner the second layer of the \textit{S-Graph}. Each of the mapped vertical planar nodes at current keyframe are further checked to assess if they belong to a corridor or a room node. The fourth module is the loop closure module which receives keyframe data from robot tracking layer to identify neighbouring keyframes and provides relative pose between the keyframes to be added as constraints within the \textit{S-Graph}.   

\subsection{Plane Extraction}
Our plane extraction module receives the raw point cloud measurements from the 3D LiDAR sensor. We first downsample the point cloud, and then remove gross outliers by filtering out points outside an interval defined by the average and standard deviation of the distances to the robot. This pre-processed point cloud is then passed to the plane segmentation module, that uses sequential RANSAC to detect all planar surfaces and gives a first estimation of their normals. In order to avoid data association errors due to planar walls having two sides close to each other, known as a double-sided issue \cite{lidar_slam_with_plane_adj}, we refer all normal orientations pointing to the LiDAR origin frame $L_t$ by converting the plane normals to its closed point form representation $\leftidx{^{L_t}}{\mathbf{\Pi}}$ as in \cite{lips}.
\begin{equation}
    \leftidx{^{L_t}}{\mathbf{\Pi}} = \leftidx{^{L_t}}{\mathbf{n}}^{\prime} \cdot \leftidx{^{L_t}}{d}^{\prime} 
\longrightarrow
    \begin{bmatrix}
     \leftidx{^{L_t}}{\mathbf{n}} \\ 
     \leftidx{^{L_t}}{{d}} 
    \end{bmatrix} =  \begin{bmatrix}
     \leftidx{^{L_t}}{\mathbf{\Pi}} / \| \leftidx{^{L_t}}{\mathbf{\Pi}} \| \\ 
     \| \leftidx{^{L_t}}{\mathbf{\Pi}} \| 
    \end{bmatrix}
\end{equation}

\noindent where $\leftidx{^{L_t}}{\mathbf{n}}=[\leftidx{^{L_t}}n_x, \ \leftidx{^{L_t}}n_y, \ \leftidx{^{L_t}}n_z]^\top$ is the plane normal and $\leftidx{^{L_t}}{d}$ is the distance to the origin, both in the LiDAR frame. 

\subsection{Robot Odometry}

We use the Voxelized Generalized Iterative Closest Point (VGICP) in \cite{vgicp}. This voxelized version of GICP aggregates the voxel distribution on each point, parallelizing the optimization and achieving similar accuracy to GICP but substantially faster (it runs at 30 Hz on a low-end CPU). Alternatively, as we run our experiments on legged robots, we also used in our \textit{S-Graphs} the odometry from the encoders of these platforms.  

\begin{figure}[]
    \centering
    \includegraphics[width=0.5\textwidth]{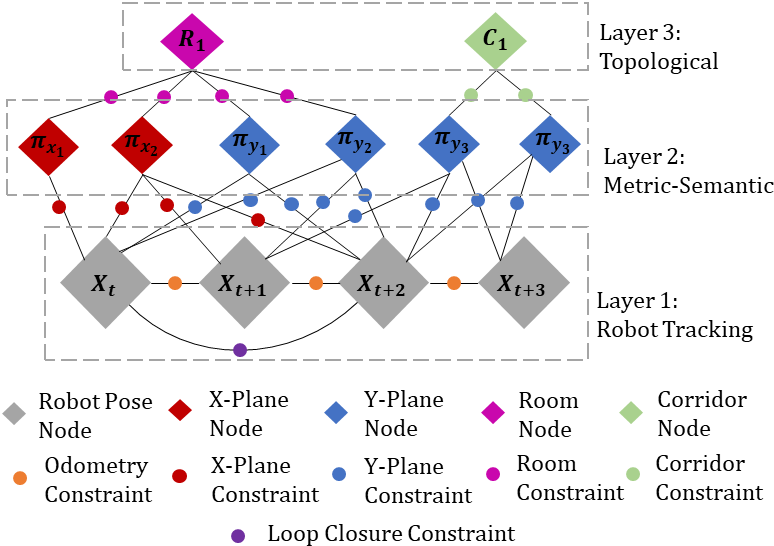}
    \caption{Example of an \textit{S-Graph} representing the robot pose, plane, room and corridor nodes together with their respective constraints.}
    \label{fig:optimizable_scene_graph}
\end{figure}

\subsection{Scene Mapping with S-Graphs} 
\label{subsec:s-graph}

This module takes as inputs the robot odometry as well as the extracted planes to create the three-layered optimizable situational graph that includes the robot states $\leftidx{^M}{\mathbf{x}}_{R_t}, \ t \in \{1, \hdots, T\}$ as well as the environmental model represented as a scene graph (see Fig.~\ref{fig:optimizable_scene_graph}). 

\textbf{Robot Tracking Layer.} (Layer 1 in Fig.~\ref{fig:optimizable_scene_graph})
This layer creates a factor node $\leftidx{^M}{\mathbf{x}}_{R_t} \in SE(3)$ with the robot keyframe pose at time $t$ in the map frame $M$. The pose nodes are constrained by pairwise odometry readings between consecutive poses $\leftidx{^{R_{t-1}}}{\tilde{\mathbf{x}}}_{R_t} \in SE(3)$. The associated cost function to minimize is
\begin{multline}
c_{\mathbf{x}}( \leftidx{^M}{\mathbf{x}}_{R_1}, \hdots, \leftidx{^M}{\mathbf{x}}_{R_T}) = \\ = \sum_{t=2}^T || \leftidx{^M}{\mathbf{x}}_{R_{t-1}}^{-1} \boxplus \leftidx{^M}{\mathbf{x}}_{R_t} \boxminus \leftidx{^{R_{t-1}}}{\tilde{\mathbf{x}}}_{R_t} ||^2_{\mathbf{\Lambda}_{\tilde{\mathbf{x}}}}, 
\end{multline}

\noindent where $\boxplus$ and $\boxminus$ are the composition and inverse composition \cite{blanco2010tutorial}, $\cdot^{-1}$ is the inverse operator, $|| \mathbf{\cdot} ||_{\mathbf{\Lambda}_{\tilde{\mathbf{x}}}}$ is the Mahalanobis distance, and $\mathbf{\Lambda}_{\mathbf{\tilde{x}}}$ is information matrix associated to $\tilde{\mathbf{x}}$.

\textbf{Metric-Semantic Layer.} (Layer 2 in Fig.~\ref{fig:optimizable_scene_graph})
This layer creates the factor nodes for the planar surfaces extracted by the planar segmentation module. The planar normals extracetd in the LiDAR frame $L_t$ at time $t$ are transformed to the global map frame $M$ for its map representation
\begin{equation}
    \begin{bmatrix}
     \leftidx{^M}{\mathbf{n}} \\ 
     \leftidx{^M}{{d}} 
    \end{bmatrix} = 
    \begin{bmatrix}
     {\leftidx{^M}{\mathbf{R}}_{L_t}} & 0 \\ 
     -\leftidx{^M}{\mathbf{t}_{L_t}} & 1 
    \end{bmatrix} 
    \begin{bmatrix}
     \leftidx{^{L_t}}{\mathbf{n}} \\ 
     \leftidx{^{L_t}}{{d}} 
    \end{bmatrix} = 
    \leftidx{^M}{\mathbf{T}}_{\boldsymbol{\pi}_t}(\leftidx{^M}{\mathbf{x}}_{R_t})\begin{bmatrix}
     \leftidx{^{L_t}}{\mathbf{n}} \\ 
     \leftidx{^{L_t}}{{d}} 
    \end{bmatrix}
\end{equation}

\noindent where we define $\leftidx{^M}{\mathbf{T}}_{\boldsymbol{\pi}_t}$ as the matrix that transforms the plane normal and distance from the LiDAR frame $L_t$ to the map frame $M$. 

The plane normals with their $\leftidx{^M}{{n}_x}$ or $\leftidx{^M}{{n}_y}$ components greater than the $\leftidx{^M}{{n}_z}$ component are classified as corresponding to vertical planes. Within the vertical planes, those with normals where $\leftidx{^M}{{n}_x}$ is greater than $\leftidx{^M}{{n}_y}$ are classified as $x$-plane normals, and otherwise they are classified as $y$-plane normals. Finally, planes whose normals' bigger component is $\leftidx{^M}{{n}_z}$ are classified as horizontal planes. 

Following \cite{cpa_slam, hdl_graph_slam} we use the minimal plane parametrization $\leftidx{^M}{\boldsymbol{\pi}} = [\leftidx{^M}\phi, \leftidx{^M}\theta, \leftidx{^M}d]$, where $\leftidx{^M}\phi$ and $\leftidx{^M}\theta$ are the azimuth and elevation of the plane in the frame $M$. The cost to minimize for each plane observation $\leftidx{^{L_t}}{\tilde{\boldsymbol{\pi}}}_i$ is as follows:
\begin{equation} \label{eq:plane_error}
    c_{\boldsymbol{\pi}}(\leftidx{^M}{\mathbf{x}}_{R_t}, \leftidx{^M}{\boldsymbol{\pi}}_i) =  \sum_{t=1, i=1}^{T, P} \| \leftidx{^{L_t}}{{\boldsymbol{\pi}}}_i - \leftidx{^{L_t}}{\tilde{\boldsymbol{\pi}}}_i \|^2_{\mathbf{\Lambda}_{\boldsymbol{\tilde{\pi}}_{i,t}}}
\end{equation}


\noindent where the predicted plane normal $\leftidx{^{L_t}}{\mathbf{n}}$ and distance $\leftidx{^{L_t}}d$ in the LiDAR frame are computed from the plane estimates and robot pose as follows
\begin{equation}
    \begin{bmatrix}
     \leftidx{^{L_t}}{\mathbf{n}} & 
     \leftidx{^{L_t}}{{d}} 
    \end{bmatrix}^\top = 
    \leftidx{^{L_t}}{\mathbf{T}}_{\boldsymbol{\pi}_t}(\leftidx{^M}{\mathbf{x}}_{R_t})\begin{bmatrix}
     \leftidx{^{M}}{\mathbf{n}} & 
     \leftidx{^{M}}{{d}}^\top 
    \end{bmatrix}
\end{equation}

\noindent and the azimuth and elevation angles in the LiDAR frame, for Eq. \ref{eq:plane_error}, are extracted from the plane normal ($\leftidx{^{L_t}}{\phi} = \atantwo(\leftidx{^{L_t}}n_y, \leftidx{^{L_t}}n_x)$ and $\leftidx{^{L_t}}{\theta} = \atantwo(\leftidx{^{L_t}}n_z, \sqrt{\leftidx{^{L_t}}n_x^2+\leftidx{^{L_t}}n_y^2})$).
After initializing each plane in the global map, correspondences are searched for every subsequent plane observations. We use the Mahalanobis distance between each mapped plane and the new extracted ones.



\textbf{Topological Layer.} \label{subsec:top_layer} (Layer 3 in Fig.~\ref{fig:optimizable_scene_graph})
The topological layer assesses if the mapped planes at a given time instance belong to a particular object or structural component and further constrain their geometry. In this work, our novelty lies in the formulation of the room and corridor nodes, which can be both represented as a set of planes, although this could be extended to other structural topologies. We define a room node as composed of four planar walls, and similarly a corridor node as composed of two parallel wall planes. To define a room node, four planar walls need to be detected and mapped at the current keyframe, and similarly a corridor node requires two detected and mapped parallel wall planes. For room nodes, if the metric-semantic layer at the keyframe node $t$ identifies four planar surfaces (two $x-$planes $\boldsymbol{\pi}_1$ and $\boldsymbol{\pi}_2$ and two $y-$planes $\boldsymbol{\pi}_3$ and $\boldsymbol{\pi}_4$), the following tests are performed:
\begin{equation}
\label{eq:room_criterion}
\begin{aligned} 
    \leftidx{^M}{\mathbf{n}_{1}} \cdot \leftidx{^M}{\mathbf{n}_{2}} < 0 &, \ \leftidx{^M}{\mathbf{n}_{3}} \cdot \leftidx{^M}{\mathbf{n}_{4}} < 0 \\
    w_x = \leftidx{^M}{{d}_{2}} - \leftidx{^M}{{d}_{1}} > \lambda &, \ w_y = \leftidx{^M}{{d}_{4}} - \leftidx{^M}{{d}_{3}} > \lambda 
\end{aligned}
\end{equation}

The dot product tests between the plane normals of the $x$-planes and the $y$-planes assess that the respective normals are opposed. $w_x$ and $w_y$ are the separation between the two $x-$planes and $y-$planes respectively, that should be greater than a threshold $\lambda$. In addition, the $x-$ and $y-$ plane pairs should be have similar extensions. 
The room node $\leftidx{^M}{\boldsymbol{\rho}} = [\leftidx{^M}\rho_x, \leftidx{^M}\rho_y, w_x, w_y]^\top$ is created using the planes satisfying the above criterion. The room center $[\leftidx{^M}\rho_x, \leftidx{^M}\rho_y]^\top$ is defined as
\begin{equation} \label{eq:room_node}
 [\rho_x, \rho_y]^\top = [(\frac{w_x}{2} + \leftidx{^M}{{d}_{1}}), (\frac{w_y}{2} + \leftidx{^M}{{d}_{3}})]^\top 
\end{equation}

Each room node is linked then to its corresponding $x-$ and $y-$ planes, and the total cost function to minimize is
\begin{equation}
c_{\boldsymbol{\rho}} = \sum_{j=1}^{S} \sum_{l=1}^{4} c_{\boldsymbol{\rho}_j, l}(\leftidx{^M}{\boldsymbol{\rho}_j}, \boldsymbol{\pi}_{l})
\end{equation}

\noindent where for each room $\leftidx{^M}{\boldsymbol{\rho}_j}$ four costs, associated to each room-plane edge, are minimized ($j$ index omitted for clarity)
\begin{equation} \label{eq:x_plane_room_node}
\begin{aligned}
    c_{\boldsymbol{\rho}, 1}(\leftidx{^M}{\boldsymbol{\rho}}, \boldsymbol{\pi}_{1}) = \| (\leftidx{^M}\rho_x - \frac{w_x}{2}) - \leftidx{^M}{\tilde{d}}_{1} \|^2_{\mathbf{\Lambda}_{\tilde{\rho}}} \\
    c_{\boldsymbol{\rho}, 2}(\leftidx{^M}{\boldsymbol{\rho}}, \boldsymbol{\pi}_{2}) = \| (\leftidx{^M}\rho_x + \frac{w_x}{2}) - \leftidx{^M}{\tilde{d}}_{2} \|^2_{\mathbf{\Lambda}_{\tilde{\rho}}} \\
    c_{\boldsymbol{\rho}, 3}(\leftidx{^M}{\boldsymbol{\rho}}, \boldsymbol{\pi}_{3}) = \| (\leftidx{^M}\rho_y - \frac{w_y}{2}) - \leftidx{^M}{\tilde{d}}_{3} \|^2_{\mathbf{\Lambda}_{\tilde{\rho}}} \\
    c_{\boldsymbol{\rho}, 4}(\leftidx{^M}{\boldsymbol{\rho}}, \boldsymbol{\pi}_{4}) = \| (\leftidx{^M}\rho_y + \frac{w_y}{2}) - \leftidx{^M}{\tilde{d}}_{4} \|^2_{\mathbf{\Lambda}_{\tilde{\rho}}} 
\end{aligned}
\end{equation}



Data association for the room node is also based on the Mahalanobis distance. We can safely tune the matching threshold close to the room widths, as rooms do not overlap. This allows us to merge planar structures duplicated due to inaccuracies. 

Similarly to rooms, corridor nodes are created from either $x$-plane or $y$-plane pairs using the criteria in Eq.~\ref{eq:room_criterion}. For $x-$corridors $\leftidx{^M}{\boldsymbol{\kappa}} = [\kappa_x, \kappa_y, w_x]$, and for $y-$corridors $\leftidx{^M}{\boldsymbol{\kappa}} = [\kappa_x, \kappa_y, w_y]$. Analogously to the room factors in Eq.~\ref{eq:room_node} and Eq.~\ref{eq:x_plane_room_node}, we formulate the corridor nodes and edges linking them to their respective planes and minimize the summation of their associated costs $c_{\boldsymbol{\kappa}} = \sum_{k=1}^{K} c_{\boldsymbol{\kappa}_k}$. Assuming $k$ corresponds to an $x-$corridor, for example, the cost to minimize would be $c_{\boldsymbol{\kappa}_k} = \sum_{l=1}^{2} c_{\boldsymbol{\kappa}_k, l}(\leftidx{^M}{\boldsymbol{\kappa}_k}, \boldsymbol{\pi}_{l})$, where  $c_{\boldsymbol{\kappa}, 1}(\leftidx{^M}{\boldsymbol{\kappa}}, \boldsymbol{\pi}_{1}) = \| (\leftidx{^M}\kappa_x - \frac{w_x}{2}) - \leftidx{^M}{\tilde{d}}_{1} \|^2_{\mathbf{\Lambda}_{\kappa}}$ and 
$\mathrm{c_{\boldsymbol{\kappa}, 2}(\leftidx{^M}{\boldsymbol{\kappa}, \boldsymbol{\pi}_{2}) = \| (\leftidx{^M}\kappa_x + \frac{w_x}{2}) - \leftidx{^M}{\tilde{d}}_{2} \|^2_{\mathbf{\Lambda}_{\kappa}}}}$. 


\subsection{Loop Closure}

In our pipeline, the loop closure is performed in a two-stage fashion. Firstly, each planar structure with its corresponding room and corridor node, adds a soft loop closure constraint optimizing the robot pose and the planar estimates mostly surrounding the room/corridor. While this soft loop constraint proves to be sufficient for environments with small areas or small robot odometry errors, an appearance-based loop closure constraint is essential in the opposite case of large odometric drifts. Hence, we also incorporate scan matching-based hard loop closure constraints modeled as edges at the robot tracking layer (see Layer 1 in Fig.~\ref{fig:optimizable_scene_graph}). The hard loop closure constraint is similar to the one implemented in \cite{hdl_graph_slam}, based on the Normal Distribution Transform-based scan matching. It uses a translational thresholding between the robot pose nodes to identify the loop closure candidates, and optimizes not only the robot poses but all the layers of the \textit{S-Graph}.

\begin{figure}[]
    \centering
    \includegraphics[width=0.45\textwidth]{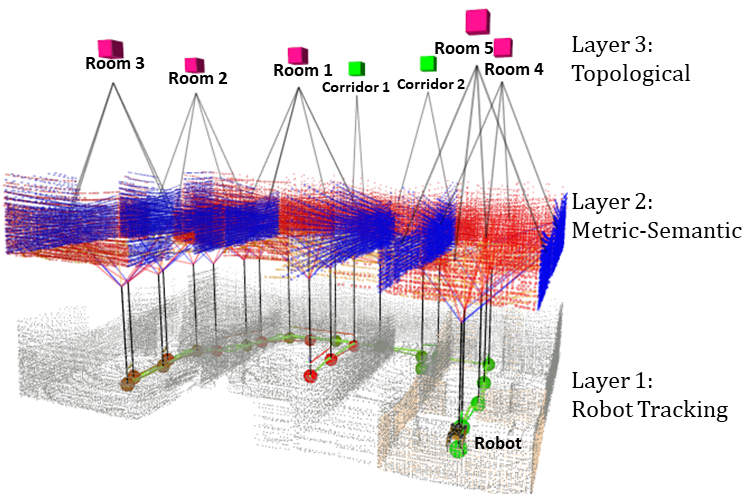}
    \caption{\textit{S-Graph} for the \textit{\mbox{SE-1}} experiment, demonstrating the creation of the three-layered situational graph. The robot nodes from the tracking layer are linked to the plane nodes from the metric-semantic layer, and plane nodes connected to the room/corridor nodes in the topological layer.}
    \label{fig:s-graph_sim}
\end{figure}
\section{Experimental Validation}

\begin{figure*}[t]
\begin{subfigure}{.5\textwidth}
\centering
\includegraphics[width=0.75\textwidth]{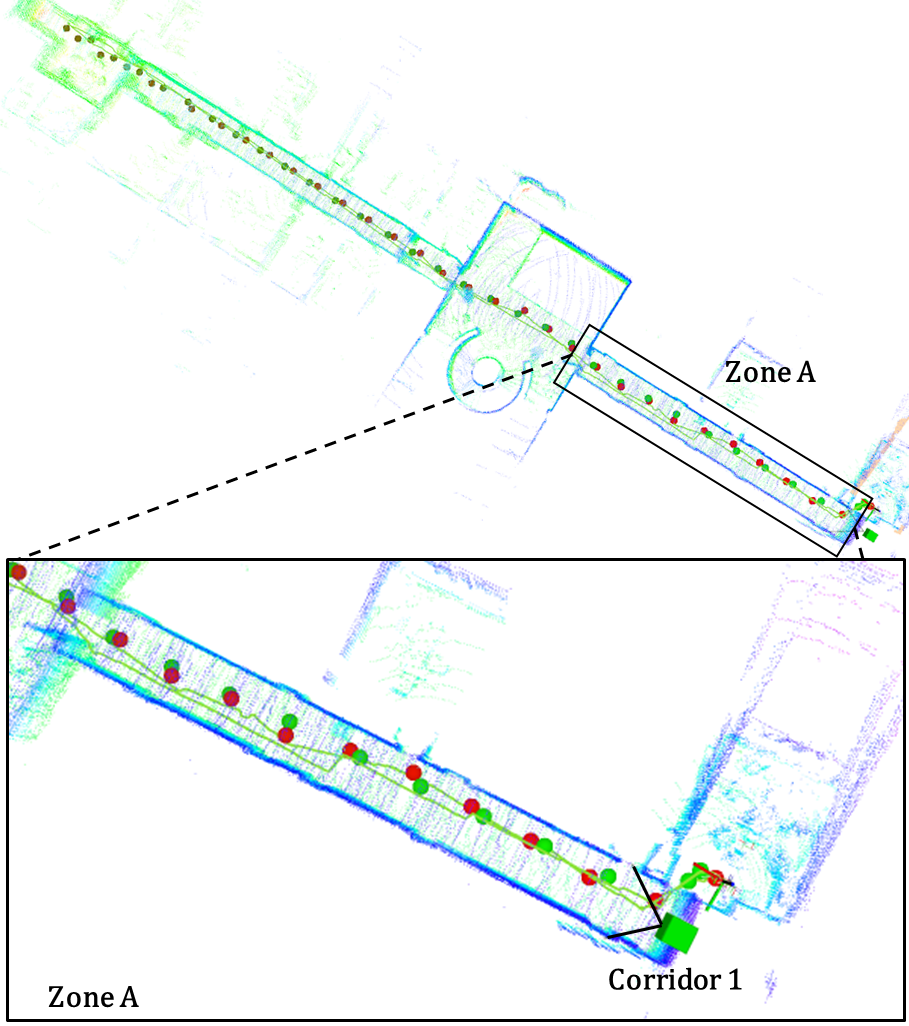}
\caption{\textit{S-Graph}}
\label{fig:}
\end{subfigure}
\rulesep
\begin{subfigure}{0.5\textwidth}
\centering
\includegraphics[width=0.75\textwidth]{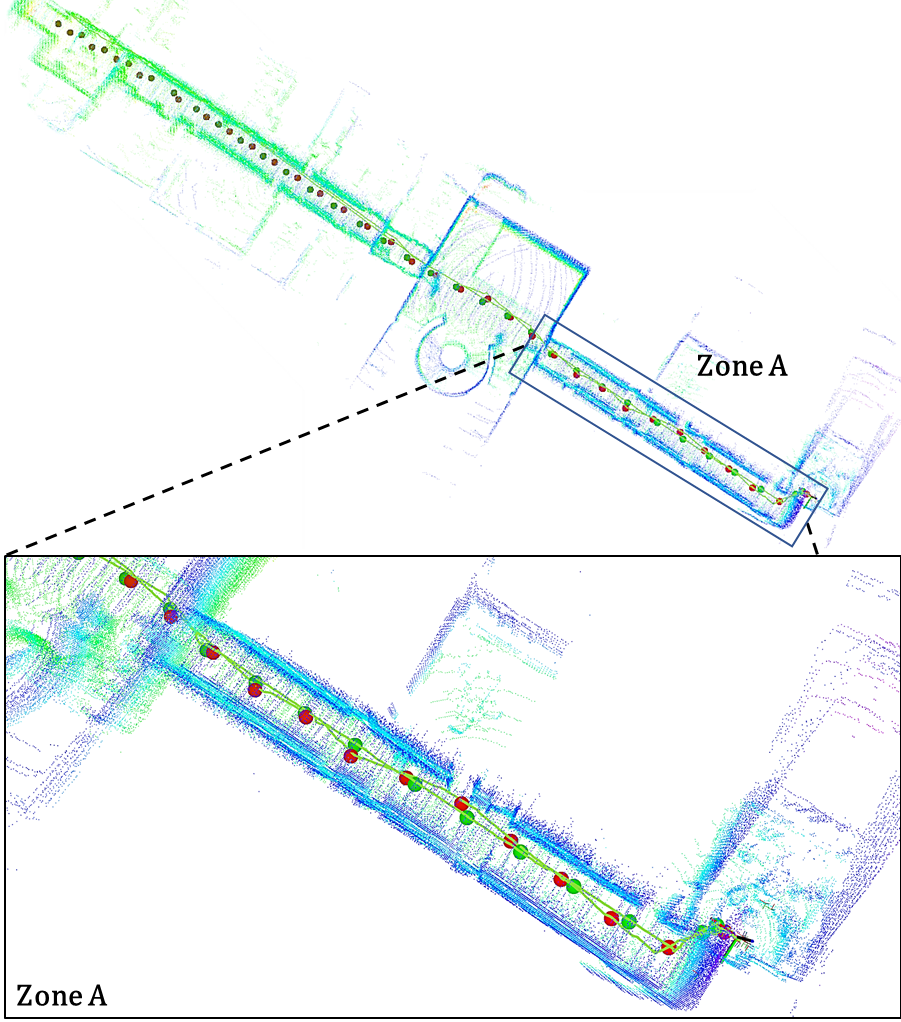}
\caption{HDL-SLAM}
\end{subfigure}
\label{fig:}
\caption{Top view of the 3D maps estimated from the real data stream \textit{LC-1} with our \textit{S-Graph} (a) and with the baseline HDL-SLAM \cite{hdl_graph_slam} (b). The zoomed-in views show the details of the mapped environment for Zone A. Note the ``cleaner'' reconstruction of the walls by our \textit{S-Graph}, indicating a more accurate alignment than HDL-SLAM.}
\label{fig:experiment_lc1}
\end{figure*}

We validate our \textit{S-Graphs} on datasets generated using both simulated and real-world indoor scenarios, comparing it against several state-of-the-art LiDAR SLAM frameworks. The datasets are collected teleoperating a Boston Dynamics Spot\footnote{\url{https://www.bostondynamics.com/products/spot}} robot equipped with a Velodyne VLP-16 3D LiDAR. \textit{S-Graph}  runs real-time on these datasets on-board an Intel i9 16~core workstation. 
Our code is built in C++ on top of the existing framework \cite{hdl_graph_slam} which only consisted of the robot tracking layer and the scan matching based loop-closure. Additional planar segmentation, metric-semantic layer and the topological layer were incorporated during this work. For identification of corridors/rooms (see Section~\ref{subsec:top_layer}) min/max length of planes to qualify as corridor candidates is $4.0$m/$15.0$m and $3.0$m/$9.0$m for room, whereas the separation thresholds between the planes is kept to $1.5$m/$3.0$m for corridors and $3.5$m/$6.0$m for rooms respectively. 
We provide both qualitative and quantitative results for the experiments. See video URL\footnote{\url{https://youtu.be/eoWrBTY04Oc}} for further results.
\subsection{Experimental Setup}
\subsubsection{Simulated Experiments}
We use the Gazebo\footnote{\url{http://gazebosim.org/}} physics simulator to re-create indoor environments. In all the simulated experiments robot encoders where not used and the odometry was estimated only from LiDAR.
We conduct a total of four simulated experiments. Two of them were generated from the 3D mesh of two floors of actual architectural plans provided by a construction company, with different configurations of walls, rooms and corridors. We denote these two settings as Construction Floor-1 (\textit{CF-1}) and Construction Floor-2 (\textit{CF-2}). We also generated two additional simulated environments resembling typical indoor environments. The first one comprises six rooms and two narrow corridors and we denote it as Simulated Environment-1 (\textit{SE-1}) (see Fig.~\ref{fig:s-graph_sim}), and the second one comprises  four bigger rooms and three corridors and we refer to it as Simulated Environment-2 (\textit{SE-2}). In all the simulated experiments, the legged robot is commanded to navigate through the environments performing several rounds and finally returning to the initial position. The simulated experiments are performed mainly to validate accuracy of the algorithms with ground truth data using Absolute Trajectory Error (ATE)\cite{evo_traj_calc} due to the absence of ground truth trajectory in real experiments.


\begin{table}[]
\centering
\caption{Absolute Trajectory Error (ATE) [m], of our \textit{S-Graph} and several baselines on simulated data. Best results are boldfaced, second best are underlined.}
\begin{tabular}{l | c c c c}
\toprule
& \multicolumn{4}{l}{\textbf{Dataset}} \\
\toprule
\textbf{Method} & \textit{CF-1} &  \textit{CF-2}  & \textit{SE-1} & \textit{SE-2}\\ \midrule
HDL-SLAM \cite{hdl_graph_slam} & 0.09 & \underline{0.11} & \underline{0.04} & \underline{0.15} \\ 
ALOAM \cite{loam} & 0.07 & 0.10 & 0.16 & 0.32 \\
MLOAM \cite{mloam} & 0.15 & 0.39 & 0.65 & 2.82 \\ 
FLOAM\cite{floam} & 3.90 & 0.44 & 0.15 & 0.24 \\ 
SCA-LOAM\cite{sca_loam} & 0.45 & 0.43 & 0.43 & 0.64 \\ 
LeGO-LOAM \cite{lego-loam}  & - & - & - & - \\
\textit{S-Graph - w/o top layer} & \underline{0.05} & {0.17} & {0.40} & {1.01}\\ 
\textit{S-Graph (ours)} & \textbf{0.04} & \textbf{0.07} & \textbf{0.03} & \textbf{0.05}\\ 
\bottomrule
\end{tabular}
\label{tab:ate_simulated_data}
\vspace{-6mm}
\end{table}


\subsubsection{Real-World Experiments}
We run a total of five experiments on different structured indoor environments ranging from construction site to office environments.
Given the presence of physical robot encoders estimating the robot odometry, we utilize the same robot encoder odometry for all the methods to have a fairer comparison. The first two experiments are performed on two floors of an on-going construction site, the same scenes whose meshes were utilized to validate the algorithm in the simulated environments (\textit{CF-1} and \textit{CF-2}). The legged robot is navigated to traverse each floor several times, to assess the capability of the methods to maintain accurate estimates of the robot poses and the 3D map. We also perform a similar experiment in an office environment with a long corridor (\textit{LC-1}) (see Fig.~\ref{fig:experiment_lc1}) that the robot traverses back and forth.  To validate the accuracy of each method on these first three experiments, we report the RMSE of the estimated 3D maps against the actual 3D map generated from the architectural plan. 


\begin{table}[ht]
\caption{Point cloud RMSE [m] on the real datasets. Best results are boldfaced, second best are underlined.}
\centering
\begin{tabular}{l | c c c}
\toprule
& \multicolumn{3}{l}{\textbf{Dataset}} \\
\toprule
\textbf{Method} & \textit{CF-1} &  \textit{CF-2} & \textit{LC-1} \\ \midrule
HDL-SLAM & 1.34 & 0.27 & \underline{1.45}  \\ 
ALOAM  & 8.03 & 1.20 & 3.14  \\ 
MLOAM  & 3.73 & 1.93 & 1.68  \\ 
FLOAM & 7.63 & 1.15 & 2.90  \\
SCA-LOAM & 4.86 & 0.75 & 2.89 \\ 
LeGO-LOAM & 4.08 & 0.70 & 3.40 \\
\textit{S-Graph - w/o top. layer} & \underline{1.21} & \textbf{0.24} & \underline{1.45}  \\ 
\textit{S-Graph (ours)} & \textbf{0.94} & \underline{0.26} & \textbf{1.32}  \\ 
\bottomrule
\end{tabular}
\label{tab:rmse_real_data}
\vspace{-3mm}
\end{table}

To assess the \textit{S-Graph} performance with loop closure constraints in the large environments, two additional experiments are performed namely \textit{LE-1} and \textit{LE-2} (Fig.~\ref{fig:experiments_le2}).~\textit{\mbox{LE-1}} comprises four corridors that the robot traverses three times. \textit{LE-2} is a similar environment comprising additional corridors that the robot traverses twice. For both experiments, since the robot is returned to the starting pose, we compute the pose error from the initial to the end point similarly to \cite{pi_slam}. A point to note in experiments \textit{LE-1} and \textit{LE-2} is that, although the robot was brought back to the initial point as carefully as possible, these metrics might be affected by errors of a few $mm$.   

\subsubsection{Ablation Study}
To evaluate the accuracy of addition of our novel room/corridor factors to the situational graph, we repeat all the experiments disabling the corridor/room factors, only enabling the robot tracking layer and the metric-semantic layer. \textit{S-Graph - w/o top. layer} can be considered a semantic SLAM framework based on planar surface mapping. This result is further analyzed in Section~\ref{subsec:ablation_study}.

\subsection{Results and Discussions}

\subsubsection{Simulated Experiments}

We compared \textit{S-Graph} against six state-of-the-art LiDAR SLAM approaches with Table~\ref{tab:ate_simulated_data} presenting the ATE of each.  
As observed in the table, our \textit{S-Graph} incorporating the entire situational graph outperforms the baseline HDL-SLAM \cite{hdl_graph_slam} by a considerable margin. \textit{S-Graph} also outperforms, by an even larger margin, other 3D LiDAR odometry and mapping systems, mainly the LOAM variant family. One of the reasons behind the higher errors of the LOAM-based variants is the use of feature-based odometries, as apposed to the VGICP-based odometry used by HDL-SLAM and our \textit{S-Graph}, which demonstrates higher accuracy. Note, however, the additional improvement given by the addition of higher-level graph levels comparing our \textit{S-Graph} against HDL-SLAM. Due to the absence of ground plane during several instances in the datasets, LeGO-LOAM \cite{lego-loam} which has high dependence on ground planes, does not provide reliable odometry results and hence neither estimates a reliable 3D map. 

\begin{figure}[t]
    \centering
    \includegraphics[width=0.5\textwidth]{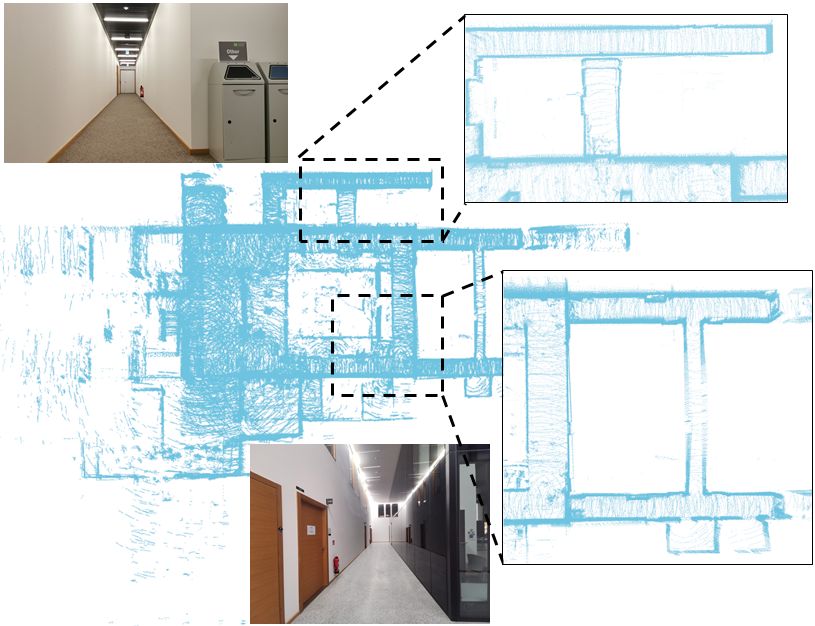}
    \caption{Top view of the 3D maps estimated using \textit{S-Graphs} during experiment \textit{LE-2}. Zoomed-in views demonstrate the high quality of the wall reconstructions (constrained using corridor nodes). We also show pictures of the zoomed areas for better understanding of the results.}
    \label{fig:experiments_le2}
\end{figure}

\subsubsection{Real Experiments} 
Table~\ref{tab:rmse_real_data} provides the RMSE of the estimated 3D maps agains the architectural plans for experiments \textit{CF-1}, \textit{CF-2} and \textit{LC-1}. As can be observed, although all the methods utilize the same odometry source, our \textit{S-Graph} which builds a higher-level topological map of the environment is able to generate accurate 3D maps of the environment compared with the other approaches. Also observe in Fig.~\ref{fig:experiment_lc1} the cleaner map generated by our \textit{S-Graph} compared to the baseline for experiment \textit{LC-1}. Semantic SLAM framework, LeGO-LOAM does not show improved performance (Table~\ref{tab:rmse_real_data}) again due to its constant dependence on ground planes, which are not present in different time instances during these experiments. For experiments \textit{LE-1} and \textit{LE-2}, Table~\ref{tab:trans_rot_err_real_data} details the translational and rotational errors. It can be observed that in from Table~\ref{tab:trans_rot_err_real_data} and Fig.~\ref{fig:experiments_le2} for experiment \textit{LE-2}, \textit{S-Graph} generates the most accurate results as it constraints the detected planes using the corridors in the environment. Experiment \textit{LE-1} contains corridors but with high presence of glass surfaces, resulting in missed corridor detections at certain time intervals eventually causing slightly higher error in the estimates of \textit{S-Graph}.

\begin{table}[]
\caption{Translational [m] and rotational [$^\circ$] errors between the initial and end points for real datasets. Best results are boldfaced, second best are underlined.}
\centering
\begin{tabular}{l | l | c c}
\toprule
& & \multicolumn{2}{l}{\textbf{Dataset}} \\
\toprule
\textbf{Method} & \textbf{Metric} & \textit{LE-1} & \textit{LE-2} \\ 
\midrule
{HDL-SLAM} & $t_{err}$ / $r_{err}$ & \textbf{0.03} / 0.17 & 0.24 / {0.06} \\ \midrule
{{ALOAM}} & $t_{err}$ / $r_{err}$ & {1.78} / {0.10} & {0.23} / 0.10 \\ \midrule
{{MLOAM}} & $t_{err}$ / $r_{err}$ & {3.60} / \textbf{0.04} & 11.2 / 0.42 \\ \midrule
{{FLOAM}} & $t_{err}$ / $r_{err}$ & {7.22} / {0.25} & 9.28 / 0.41 \\ \midrule
{{SCA-LOAM}} & $t_{err}$ / $r_{err}$ & {0.55} / \underline{0.08} & 0.93 / 0.09 \\  \midrule
{{LeGO-LOAM}} & $t_{err}$ / $r_{err}$ & 0.32 / 0.10 & \underline{0.15} / 0.72 \\  \midrule
{\textit{S-Graph - w/o top. layer}} & $t_{err}$ / $r_{err}$ & {0.49} / {0.12} & {0.32} / \underline{0.05} \\ \midrule
{\textit{S-Graph (ours)}} & $t_{err}$ / $r_{err}$ & \underline{0.08} / 0.14 & \textbf{0.11} / \textbf{0.04} \\
\bottomrule
\end{tabular}
\vspace{-3mm}
\label{tab:trans_rot_err_real_data}
\end{table}

\subsubsection{Ablation Study} \label{subsec:ablation_study}

Table~\ref{tab:ate_simulated_data} presents the results in simulated datasets without the topological layer. It can be clearly seen the advantage of adding our novel room/corridor factors, in particular in the larger scenes of experiments \textit{SE-1} and \mbox{\textit{SE-2}}. In these two cases, the odometry drift causes erroneous plane matches, that are  corrected due to the addition of the topological layer (explained in Section~\ref{subsec:top_layer}). Although, \textit{S-Graph - w/o top. layer} in Table~\ref{tab:rmse_real_data} presents slightly higher to equal errors in point cloud alignment in the absence of the topological layer, observe the higher error in larger environments in Table~\ref{tab:trans_rot_err_real_data}.

\subsubsection{Limitations}
Our \textit{S-Graph} approach has shown promising results over all evaluated datasets, maintaining real-time performance for datasets with sequence lengths over $26$ mins (see Table~\ref{tab:compute_time}). However, note the larger computation times of the map optimizations compared to the baselines HDL-SLAM and ALOAM, which is expected due to the additional planar and room/corridor nodes and their constraints. Also note high standard deviation in some cases as time variations depend strongly on size and structure of the graph. We consider to address these limitations in future work by optimizing only the \textit{S-Graph} nodes of the neighbouring region to the room/corridor that the robot is currently in. 

\begin{table}[h]
\caption{Computation time [ms] of the most promising methods along with the total length of the sequence [s] of each real dataset.}
\centering
\begin{tabular}{l l | l l l}
\toprule
& & \multicolumn{3}{l}{\textbf{Computation Time} (mean/std) [ms]} \\ 
\toprule
\textbf{Dataset} & \textbf{\vtop{\hbox{\strut Sequence}\hbox{\strut Length}}} [s] & {HDL-SLAM} &  ALOAM & \textit{S-Graph (ours)} \\ \midrule
\textit{CF-1} & 487 & 3.9 / 4.2 & 3.3 / 2.1  & 70 / 89.1 \\
\textit{CF-2} & 657  & 3.4 / 3.3 & 2.2 / 1.6 & 83.8 / 85.3  \\ 
\textit{LC-1} & 339  & 8.1 / 7.7 & 3.6 / 1.9 & 80.1 / 104.9 \\ 
\textit{LE-1} & 1321 & 22.5 / 19.4 & 5.1 / 1.8 & 257.7 / 447.3 \\
\textit{LE-2} & 1585 & 10.3 / 10.8 & 5.2 / 3.6 & 330.6 / 396.1 \\
\bottomrule
\end{tabular}
\label{tab:compute_time}
\vspace{-5mm}
\end{table}
\section{Conclusion}

In this paper we present \textit{S-Graphs}, Situational Graphs for robots in structured environments, bridging the gap between low-level geometric/semantic SLAM and recent approaches targeting higher-level scene graphs. Our \textit{S-Graphs} are composed of three layers. The \textit{Robot Tracking Layer} optimizes a set of keyframe poses using odometric constraints. The \textit{Metric-Semantic Layer} creates a dense metric-semantic map of the environment composed of planar surfaces, associating each plane with the keyframes in the robot tracking layer where they are visible. The \textit{Topological Layer}, as the highest-level layer of the graph, connects the mapped planar structures with our novel room/corridors factors. 

We validated \textit{S-Graphs} on simulated and real datasets captured by a legged robot in a construction site as well as in large structured indoor environments. We compared our approach against several relevant LiDAR-SLAM baselines, outperforming the current state-of-the-art. This showcases that incorporating environment representations in the form of  hierarchical \textit{S-Graphs} do not only enhances the understanding of the environment, but also improves the state estimates. In addition to the future lines mentioned above, we plan to incorporate additional structural and dynamic constraints into \textit{S-Graphs}, validating it over non-manhattan worlds and improve the room/corridor extraction and matching using related works referred in \cite{roomsegmentationreview}, \cite{3d_room_recontruction}

\balance
\bibliographystyle{IEEEtran}
\bibliography{Bibliography}

\vfill

\end{document}